\documentclass[11pt,a4paper]{article}
\usepackage{times,latexsym}
\usepackage{url}
\usepackage[T1]{fontenc}

\usepackage[acceptedWithA]{tacl2021v1}

\usepackage{xspace,mfirstuc,tabulary}

\newif\iftaclinstructions
\taclinstructionsfalse 
\iftaclinstructions
\renewcommand{\confidential}{}
\renewcommand{\anonsubtext}{(No author info supplied here, for consistency with
TACL-submission anonymization requirements)}
\newcommand{\instr}
\fi

\iftaclpubformat 

\else

\fi

\usepackage{booktabs}
\usepackage{xcolor}
\usepackage{colortbl}
\usepackage{amsmath}
\usepackage{multirow}
\usepackage{graphicx}
\usepackage{float}
\usepackage{subcaption}
\usepackage{cleveref}
\usepackage{booktabs,arydshln}

\makeatletter
\def\adl@drawiv#1#2#3{%
        \hskip.5\tabcolsep
        \xleaders#3{#2.5\@tempdimb #1{1}#2.5\@tempdimb}%
                #2\z@ plus1fil minus1fil\relax
        \hskip.5\tabcolsep}
\newcommand{\cdashlinelr}[1]{%
  \noalign{\vskip\aboverulesep
           \global\let\@dashdrawstore\adl@draw
           \global\let\adl@draw\adl@drawiv}
  \cdashline{#1}
  \noalign{\global\let\adl@draw\@dashdrawstore
           \vskip\belowrulesep}}
\makeatother

\usepackage{lipsum}

\newcommand{\keypoint}[1]{\noindent \textbf{#1}:}

\usepackage{todonotes}

\date{}

\begin{document}


\title{CG-TTRL: Context-guided Test-time Reinforcement Learning for On-device Large Language Models}



\author{
  \textbf{Peyman Hosseini\textsuperscript{1,2}} \Thanks{Work done during internship at Samsung R\&D Institute UK.} \quad
  \textbf{Ondrej Bohdal\textsuperscript{1}} \quad
  \textbf{Taha Ceritli\textsuperscript{1}} \quad
  \textbf{Ignacio Castro\textsuperscript{2}} \\
  \textbf{Matthew Purver\textsuperscript{2}} \quad
  \textbf{Mete Ozay\textsuperscript{1}} \quad
  \textbf{Umberto Michieli\textsuperscript{1}}
  \\[0.5em]
  \textsuperscript{1}Samsung R\&D Institute UK \quad
  \textsuperscript{2}Queen Mary University of London \\
  {\small \texttt{\{s.hosseini, i.castro, m.purver\}@qmul.ac.uk}} \\
  {\small \texttt{\{o.bohdal.1, t.ceritli, m.ozay, u.michieli\}@samsung.com}
}}




\maketitle
\begin{abstract}
Test-time Reinforcement Learning (TTRL) has shown promise in adapting foundation models for complex tasks at test-time, resulting in large performance improvements. TTRL leverages an elegant two-phase sampling strategy: first, multi-sampling derives a pseudo-label via majority voting, while subsequent downsampling and reward-based fine-tuning encourages the model to explore and learn diverse valid solutions, with the pseudo-label modulating the reward signal. Meanwhile, in-context learning has been widely explored at inference time and demonstrated the ability to enhance model performance without weight updates. However, TTRL's two-phase sampling strategy under-utilizes contextual guidance, which can potentially improve pseudo-label accuracy in the initial exploitation phase while regulating exploration in the second. To address this, we propose context-guided TTRL (CG-TTRL), integrating context dynamically into both sampling phases and propose a method for efficient context selection for on-device applications. Our evaluations on mathematical and scientific QA benchmarks show CG-TTRL outperforms TTRL (e.g. additional 7\% relative accuracy improvement over TTRL), while boosting efficiency by obtaining strong performance after only a few steps of test-time training (e.g. 8\% relative improvement rather than 1\% over TTRL after 3 steps).
\end{abstract}

\section{Introduction}
\label{sec: Introduction}
Personalizing foundation models to distribution shifts and domain-specific tasks (e.g., mathematical reasoning) with minimal computational overhead has become a popular area for Artificial Intelligence (AI) research. Traditional approaches to training language models often rely on static datasets and pre-defined objectives, limiting their adaptability to new or complex tasks \citep{Berglund+24}. Test-Time Adaptation (TTA) addresses this challenge by enabling models to dynamically adjust to new tasks during inference \citep{Sun+20,Niu+22,Zhou+25}. In recent years, the integration of reinforcement learning (RL) with language models has gained significant attention for its potential to enhance reasoning and decision-making capabilities at test time \citep{Ouyang+22,Zuo+25}. 

Lying within the aforementioned paradigms, Test-Time Reinforcement Learning (TTRL) \citep{Zuo+25}, aims to dynamically adapt to new tasks at test-time without any external supervision. It refines model outputs through iterative interactions with feedback mechanisms. However, TTRL faces challenges such as reliance on noisy reward signals and limited generalization, a common problem in RL-based algorithms \citep{JiangKR23}, hindering its adaptability in practice. Provision of context both in traditional RL setups such as in Robotics \cite{Benjamin+23} and in new RL applications in AI agents \cite{Lee+23} has shown to significantly improve the performance of these approaches. The definition of context is broad in such settings. It can vary from additional information about the environment such as arm stiffness or gravity in Robotics \citep{Benjamin+23} or similar question-answer pairs in novel applications of RL where foundation models act as AI agents \citep{Lee+23}. The latter is widely explored by the Natural Language Processing (NLP) community under the name of In-Context Learning (ICL).

In-Context Learning \citep{Brown+20} has been increasingly adopted to improve the performance of Large Language Models (LLMs) in complex tasks such as Mathematical Reasoning \citep{Zhou+22,Agarwal+24}, Code Generation \citep{Patel+24}, and Medical Diagnostics \citep{Ferber+24} without altering model weights at inference time. Additionally, the context window of foundation models has been continuously expanding, with recent open-source models having context windows surpassing tens of thousands of tokens \citep{Qwen3+25,Phi4+25} or even hundreds of thousands of tokens \citep{Deepseek+24,Gemma3+25,Llama4+25}. These advancements enable inclusion of a large number of context examples or task demonstrations in the query. At the same time, previous research has shown unrepresentative or misleading examples can even cause performance degradation \citep{Zhao+21,Min+22,YeD22}. Therefore, increasing the number of context examples alone does not usually lead to performance improvement.

Recent research has focused on advancing context selection algorithms \citep{Dong+24,Mao+25}, employing techniques such as contrastive learning \citep{GaoD24,Zhang+24} or iterative filtering \citep{LiQ23} to identify high-impact examples. However, these methods often rely on auxiliary models or computationally intensive metrics, limiting their utility in resource-constrained settings. For Test-Time Adaptation (TTA)—which requires dynamic context selection to guide learning—and on-device LLMs, such overhead is impractical. Lightweight, storage-efficient algorithms are thus essential. Efficiency is critical for personalization tasks, where context must dynamically adapt to user patterns or domain shifts. Current techniques prioritize accuracy over efficiency, neglecting edge deployment trade-offs. Our work bridges this gap with lightweight context selection mechanisms tailored for TTRL, ensuring scalable adaptation.

\begin{figure}[htbp]
    \centering
    \includegraphics[width=0.85\columnwidth]{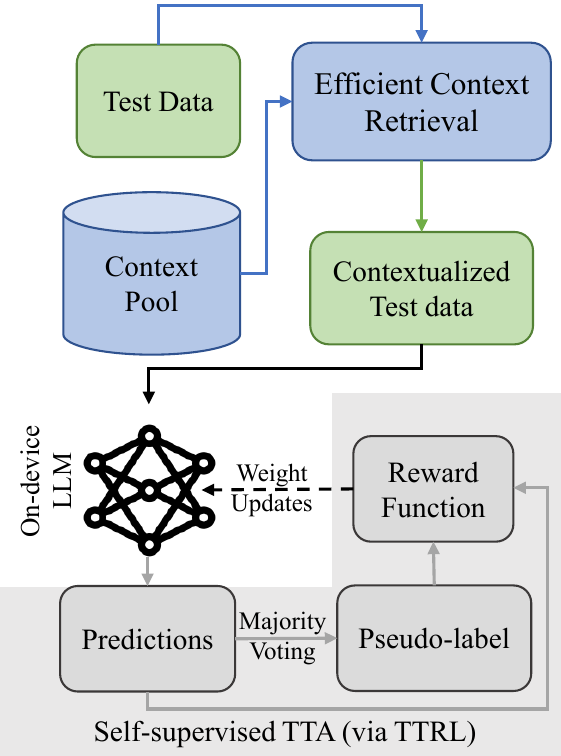}
    \caption{Our CG-TTRL approach improves TTRL by efficiently selecting context.}
    \label{fig:overview}
\end{figure}

\paragraph{Contribution:} Our paper's contributions can be summarized as follows:
\begin{itemize}
    \item CG-TTRL framework (Figure~\ref{fig:overview}): To our knowledge, we propose the first framework to integrate in-context learning (ICL) into an unsupervised fine-tuning pipeline, specifically enhancing TTRL’s two-phase strategy. This approach dynamically leverages context to refine pseudo-labels during the exploitation phase (i.e., majority voting after multi-sampling) and regulate exploration phase (i.e., fine-tuning on diverse explored solutions leading to the pseudo-label), addressing a gap in self-supervised adaptation.
    \item Lightweight context selection: We design efficient context selection methods tailored for on-device deployment, prioritizing storage and computational efficiency without compromising adaptation quality.
    \item Empirical validation: Through evaluations on mathematical and scientific QA benchmarks, we demonstrate CG-TTRL’s superior accuracy and fast performance improvement compared to TTRL, even under low-resource constraints.
    \item Edge-device scalability: Our work establishes a paradigm for continuous, context-aware self-improvement, addressing the trade-off between efficiency and personalization for edge-device applications.
\end{itemize}

\section{Related Work}
\label{sec: Related Work}
\paragraph{Test-Time Adaptation:} Test-Time Adaptation (TTA) enables models to adjust to distribution shifts during inference. Self-supervised methods like entropy minimization \citep{Wang+21,GaoZL24} and pseudo-labeling \citep{Goyal+22,Yu+24} refine predictions without labeled data. Test-Time Reinforcement Learning (TTRL) \citep{Zuo+25} uses a two-phase sampling strategy to balance exploitation and exploration, computing rewards that are used to fine-tune the model. However, TTRL's reliance on noisy pseudo-labels and limited contextual guidance restricts its robustness to various types of questions. Integrating contextual cues--inspired by robotics \citep{Benjamin+23} and AI agents \citep{Lee+23}--has shown promise in stabilizing adaptation of models. Our CG-TTRL framework addresses these gaps by dynamically incorporating context to refine pseudo-labels and regulate exploration enhancing both accuracy and performance improvement speed for on-device applications.

\paragraph{Context Selection:} In-Context Learning (ICL) relies on selecting representative examples to guide model predictions. Early approaches used random selection \citep{Brown+20}, while later methods employed semantic similarity via embeddings \citep{Liu+22} or retrieval-augmented generation \citep{Karpukhin+20}. Recent techniques leverage contrastive learning \citep{GaoD24} and iterative filtering \citep{LiQ23} to identify high-impact examples, but these often require auxiliary models or intensive computations. For edge deployment, lightweight methods are critical. Sparse attention mechanisms \citep{Child+19} and locally-sensitive hashing \citep{KitaevKL20} reduce computational overhead, while storage-efficient algorithms such as coreset selection \citep{Coleman+19}  optimize context retention. Our proposed context selection approach prioritizes efficiency and feasibility using minimal compute and storage resources to dynamically curate relevant examples for continuous adaptation at test time.

\paragraph{On-device LLMs:} Deploying LLMs on edge devices necessitates addressing computational and memory constraints, particularly for dynamic adaptation scenarios like Test-Time Adaptation (TTA). While traditional LLMs require high-end hardware for inference \citep{Borzunov+23}, on-device LLMs such as Llama 3.2 1B \citep{Llama3+24}, Gemma 3 1B \citep{Gemma3+25}, and Qwen2.5 1.5B \citep{Qwen2.5+24} offer a balance between performance and efficiency, enabling localized processing of sensitive data \citep{Dhar+21}. These models leverage compression techniques to reduce footprint, yet face trade-offs in adaptability when handling distribution shifts. As a standard practice, such foundation models are adapted for downstream tasks by adapters deployed on the device \citep{Apple+24,Dong+24}. Adapter merging is common to support multi-tasking and save storage \cite{bohdal2025device, ceritli2025hydraopt, shenaj2025k}.

\section{Methodology}
\label{sec: Methodology}
\subsection{Setup}

We first summarize test-time reinforcement learning, which we extend and improve in our work. As part of TTRL, the user provides a query $q$, which often corresponds to a challenging reasoning problem, for which TTA is helpful. TTRL generates $M$ predictions $\{\hat{y}_i\}_{i=1}^{M}$ to identify a pseudo label $y$ via majority voting. The next step is to compute a reward $R$ that can be used to fine-tune the model with parameters $\theta$. To compute the reward, $N<M$ already generated samples $\{\hat{y}^\prime_i\}_{i=1}^{N}$ are selected and compared with the pseudo label $y$. If the generated sample $\hat{y}^\prime_i$ is the same as the pseudo label $y$, the reward $R\left(\hat{y}^\prime_i, y\right)$ becomes 1, otherwise 0. The objective is to maximize the expected reward by optimizing the parameters $\theta$ via gradient ascent on the computed reward.

\subsection{Context-guided TTRL}

Our primary goal is to improve the performance of TTRL, while ensuring suitability for on-device use cases. Such scenarios require solutions that are efficient in terms of both storage and runtime. We design a simple technique that is computationally lightweight, does not require storing additional models and most importantly leads to significant improvements over the performance of TTRL.

\begin{figure*}[htbp]
    \centering
    \includegraphics[width=1\textwidth]{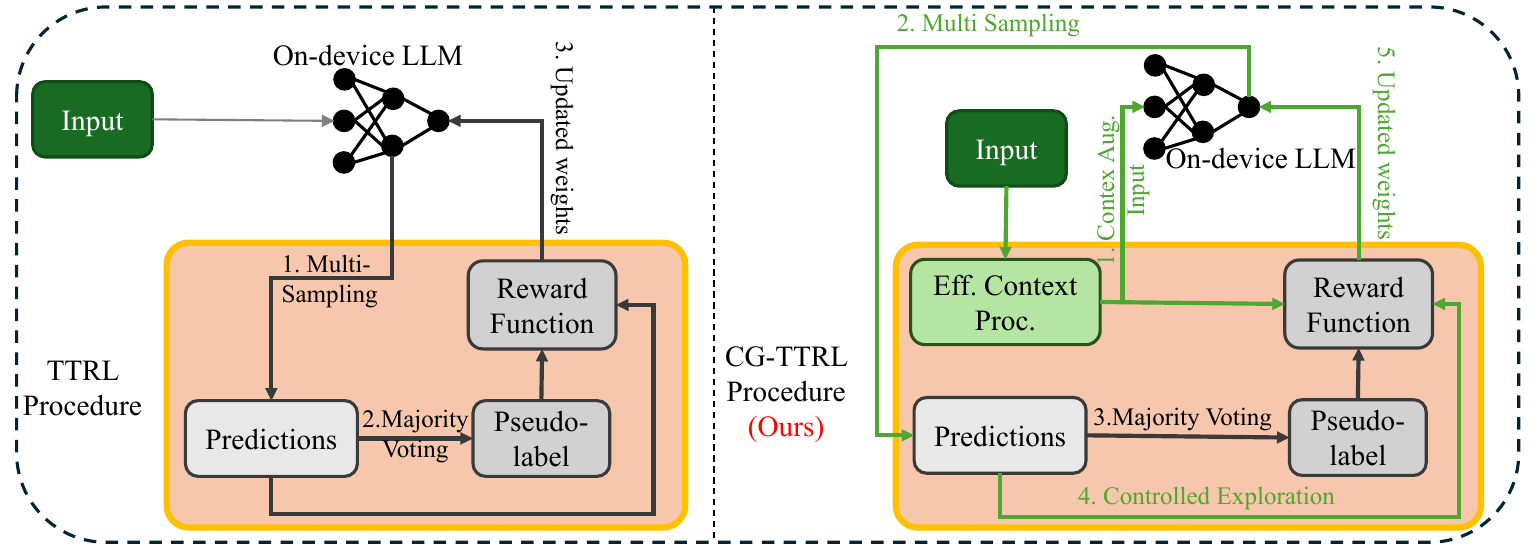}
    \caption{Comparison of vanilla TTRL (left) and our CG-TTRL that efficiently adds context (right).}
    \label{fig:comparison}
\end{figure*}

We propose to extend TTRL via efficiently-selected context to improve its performance, and as it turns out, also the ability to obtain strong performance improvement quickly. Figure~\ref{fig:comparison} provides a high-level comparison between our context-guided TTRL (CG-TTRL) and vanilla TTRL. We include $C$ context examples $\{q^c_i, s^c_i, y^c_i\}_{i=1}^C$, where $q^c_i$ denotes the example queries, with detailed step-by-step solutions $s_i^c$ and final responses $y^c_i$. We use these examples to extend the query $q$ as $q^\prime=\{q^c_i, s^c_i, y^c_i\}_{i=1}^C \cup q$.

There may be a high number of  examples relevant to a query. Therefore, we develop an efficient method to identify the most promising ones. The key idea of our solution is to utilize TF-IDF features to efficiently identify the most useful examples. TF-IDF score \cite{jones1988tfidf} consists of two components, the term frequency $\text{TF}(t,d)$ and inverse document frequency $\text{IDF}(t)$ for term $t$ and document $d$. $\text{TF}(t,d)$ is computed as the number of times term $t$ appears in document $d$ out of the total number of terms in document $d$. $\text{IDF}(t)$ is computed as the logarithm of the total number of documents in the corpus over the number of documents containing term $t$. In our case the corpus corresponds to all available examples in the context pool and document corresponds to a particular example that can be used within the context. $\text{TF-IDF}$ score for term $t$ and document $d$ is then simply computed as $\text{TF}(t,d)\times\text{IDF}(t)$. We only extract the TF-IDF features for the query components and not for the responses or detailed step-by-step solutions.

We assume that the available examples are stored as pairs of text (query, detailed solution and final response) and TF-IDF features for the query part of the text. When the user gives a new query $q$, we perform the following steps: 1) extract the TF-IDF features for query $q$: $f_{TF-IDF}(q)$, 2) find the most similar examples in the context pool in terms of TF-IDF features, 3) select top $C$ examples in terms of similarity to include as context in the prompt.

We compute the similarity between query $q$ and all $K$ available queries $\{q^p_i\}_{i=1}^K$ in the context pool using cosine similarity as:
$$\text{sim}(q,q_i^p)=\cos(f_{TF-IDF}(q), f_{TF-IDF}(q_i^p)).$$

\begin{figure}[htbp]
    \centering
    \includegraphics[width=1\columnwidth]{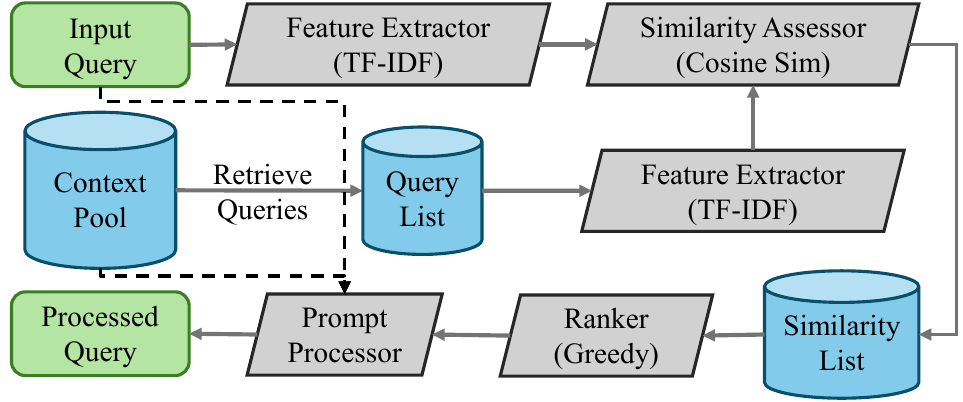}
    \caption{Our CG-TTRL approach efficiently finds the most similar queries in the context pool and uses them to modify the prompt.}
    \label{fig:context_selection_overview}
\end{figure}

We then order the available queries in terms of the similarity and select the $C$ queries with the largest similarity to query $q$. These queries from the overall context pool and their associated step-by-step solutions as well as final responses will form the context set $\{q^c_i, s^c_i, y^c_i\}_{i=1}^C$ that is combined with query $q$. Figure~\ref{fig:context_selection_overview} presents an overview of the described context selection mechanism and how it is used within the overall CG-TTRL approach.

TF-IDF is a lightweight way to extract compact features as it only consists of a relatively small number of counting operations and does not require any additional model that would need to be stored on the device. At the same time, as we show in an analysis within our experimental evaluation, its performance is competitive with more complex and significantly less efficient variations. Overall our approach leads to an efficient yet well-performing way to improve TTRL in on-device applications.

\section{Evaluation}
\label{sec: Evaluation}
\subsection{Evaluation Details}

\keypoint{Datasets} We have tested and analysed our solution on a number of challenging reasoning benchmarks, in particular MATH-500 \citep{MATH-500}, GSM-100 \citep{GSM-8k}, AMC \citep{AMC}, AIME 2024 \citep{AIME-2024} and GPQA \citep{GPQA}, following the paper introducing TTRL \cite{Zuo+25}. MATH-500, GSM-100, AMC, AIME are mathematical reasoning benchmarks, while GPQA is a challenging Q\&A benchmark that includes questions from other domains. We use MATH-500 and AMC datasets for fine-tuning the model (only one dataset is used for fine-tuning in one experiment) and also consider out-of-distribution evaluation on the remaining datasets. Evaluation on GPQA corresponds to a case of a large domain shift, while the other mathematical reasoning benchmarks correspond to a milder domain shift as the fine-tuning is done on a mathematical reasoning benchmark. The performance is evaluated in terms of accuracy on the test samples.
Context examples for GSM-100 and MATH-500 come from within these datasets. GSM-100 includes a set of 15 examples with detailed solutions, so we use these as the pool for selecting in-context examples. MATH-500 includes step-by-step solutions for all 500 examples. We consider all other examples than the current test one as the context pool.
For the other three datasets (i.e., AMC, AIME, and GPQA), the datasets only come with the question and ground-truth answer and do not include a step-by-step solution. To address this, we consider the MATH-500 dataset as the context pool as it is the largest and most diverse (containing questions of different domains and difficulty levels in mathematics). Hence we search for similar context examples from the MATH-500 dataset.

\keypoint{Models} We focus on models suitable for on-device deployment, in particular models with 1.5B parameters. For our main experiments we use Qwen2.5-Math-1.5B-Instruct \citep{Qwen2.5-Math} as an example of model specialized for mathematical reasoning, and Qwen2.5-1.5B-Instruct \citep{Qwen2.5+24,Qwen2.5+25} and DeepSeek-R1-Distill (1.5B) \citep{DeepSeek-R1} models as examples of more general-purpose models. For additional analyses we also consider Qwen2.5-Math-7B-Instruct \citep{Qwen2.5-Math} to study scaling to larger models, and Qwen2.5-Math-1.5B \citep{Qwen2.5-Math} to study the impact of using models that are not instruction-tuned.

\keypoint{Baselines} We compare with a vanilla zero-shot approach and with TTRL, which we extend as part of our CG-TTRL solution.

\keypoint{Hyperparameters} We select the hyperparameters and other details for CG-TTRL training and evaluation based on performance on validation data. In particular, based on validation data performance we select $C=3$ as the number of context set examples when using MATH-500 dataset and $C=5$ when using AMC dataset. Fine-tuning is done for 40 epochs (120 steps) on MATH-500 dataset, and 50 epochs (100 steps) for  AMC, with Adam optimizer and a learning rate of $5 \times 10^{-7}$ for the actor model and $9 \times 10^{-6}$ for the critic model consistent with the original TTRL paper. The maximum input and generation length is set to 2048 tokens, taking into account limitations of on-device language models. As part of TTRL, we use temperature $\tau=1$ to achieve a balance between exploitation and exploration characteristics of the algorithm consistent with the TTRL paper. We set $\tau=0$ for the final stage, where we evaluate the performance of the models after fine-tuning. We sample $M=32$ (64 for the AMC dataset) outputs for the majority-voting stage and downsample $N=16$ for the reward calculation and fine-tuning stage. We used 2$\times$NVIDIA A100 80GB GPUs for each experiment.


\subsection{Experimental Results}

\keypoint{Main Results} We report the main results in Table~\ref{tab:ttrl_results_main} (bold results highlight the best performance for each model).
We use either MATH-500 or AMC datasets for test-time training (TTT) and evaluate on queries from all considered datasets. Hence we include evaluation on both in-domain and out-domain examples, with the out-of-domain examples coming either from the other mathematical reasoning benchmarks or GPQA that represents a more substantial domain shift to non-mathematical reasoning. The results indicate strong improvements compared to both TTRL and the zero-shot approach where we use the model directly without any TTA. For example, DeepSeek-R1-Distill has on average performance of 37.2\% without TTA, TTRL boosts the perfomance to 56.8\% (relative improvements of 52.7\%) and our CG-TTRL improves the performance further to 59.4\% (relative improvement of 59.7\%).

\begin{table*}[ht]
\centering
\label{tab:ttrl_results_main}
\resizebox{\textwidth}{!}{%
\begin{tabular}{llcccccc}
\toprule
& \textbf{Name} & \textbf{MATH-500} & \textbf{GSM-100} & \textbf{AMC} & \textbf{AIME 2024} & \textbf{GPQA} & \textbf{Avg.} \\
\midrule
\parbox[t]{2mm}{\multirow{18}{*}{\rotatebox{90}{\textbf{TTT on MATH-500}}}} 
& Qwen2.5-Math-1.5B-Instruct & 72.8\% & 86.1\% & 48.2\% & 10.0\% & 27.3\% & 48.9\% \\
& \cellcolor{blue!5} w/ TTRL & \cellcolor{blue!5} 78.6\% & \cellcolor{blue!5} 91.7\% & \cellcolor{blue!5} 54.2\% & \cellcolor{blue!5} 20.0\% & \cellcolor{blue!5} 37.4\% & \cellcolor{blue!5} 56.4\% \\
& \cellcolor{blue!5} Rel. $\Delta_{\text{TTRL}/\text{Base}}$ & \cellcolor{blue!5} \textcolor{red}{+8.0\%} & \cellcolor{blue!5} \textcolor{red}{+6.5\%} & \cellcolor{blue!5} \textcolor{red}{+12.4\%} & \cellcolor{blue!5} \textcolor{red}{+100\%} & \cellcolor{blue!5} \textcolor{red}{+37.0\%} & \cellcolor{blue!5} \textcolor{red}{+15.3\%} \\

\cdashlinelr{2-8} 

& \cellcolor{green!5} w/ CG-TTRL & \cellcolor{green!5} \textbf{79.2}\% & \cellcolor{green!5} \textbf{94.4}\% & \cellcolor{green!5} \textbf{56.6}\% & \cellcolor{green!5} \textbf{23.3}\% & \cellcolor{green!5} \textbf{38.4}\% & \cellcolor{green!5} \textbf{58.4}\% \\
& \cellcolor{green!5} Rel. $\Delta_{\text{CG-TTRL}/\text{Base}}$ & \cellcolor{green!5} \textcolor{red}{+8.8\%} & \cellcolor{green!5} \textcolor{red}{+9.6\%} & \cellcolor{green!5} \textcolor{red}{+17.4\%} & \cellcolor{green!5} \textcolor{red}{+133.3\%} & \cellcolor{green!5} \textcolor{red}{+40.7\%} & \cellcolor{green!5} \textcolor{red}{+19.4\%} \\

\cmidrule{2-8} 
& Qwen2.5-1.5B-Instruct & 54.6\% & 72.2\% & 21.7\% & 3.3\% & 28.3\% & 36.0\% \\

& \cellcolor{blue!5} w/ TTRL & \cellcolor{blue!5} 62.6\% & \cellcolor{blue!5} 83.3\% & \cellcolor{blue!5} 32.5\% & \cellcolor{blue!5} \textbf{13.3}\% & \cellcolor{blue!5} 34.3\% & \cellcolor{blue!5} 45.2\% \\
& \cellcolor{blue!5} Rel. $\Delta_{\text{TTRL}/\text{Base}}$ & \cellcolor{blue!5} \textcolor{red}{+14.6\%} & \cellcolor{blue!5} \textcolor{red}{+15.4\%} & \cellcolor{blue!5} \textcolor{red}{+49.8\%} & \cellcolor{blue!5} \textcolor{red}{+303.0\%} & \cellcolor{blue!5} \textcolor{red}{+21.2\%} & \cellcolor{blue!5} \textcolor{red}{+25.6\%} \\

\cdashlinelr{2-8} 

& \cellcolor{green!5} w/ CG-TTRL & \cellcolor{green!5} \textbf{64.0}\% & \cellcolor{green!5} \textbf{86.1}\% & \cellcolor{green!5} \textbf{36.1}\% & \cellcolor{green!5} 10.0\% & \cellcolor{green!5} \textbf{34.8}\% & \cellcolor{green!5} \textbf{46.2}\% \\
& \cellcolor{green!5} Rel. $\Delta_{\text{CG-TTRL}/\text{Base}}$ & \cellcolor{green!5} \textcolor{red}{+17.2\%} & \cellcolor{green!5} \textcolor{red}{+19.3\%} & \cellcolor{green!5} \textcolor{red}{+66.4\%} & \cellcolor{green!5} \textcolor{red}{+203.0\%} & \cellcolor{green!5} \textcolor{red}{23.0\%} & \cellcolor{green!5} \textcolor{red}{+28.3\%} \\

\cmidrule{2-8} 

& DeepSeek-R1-Distill & 51.0\% & 77.8\% & 22.9\% & 3.3\% & 30.8\% & 37.2\% \\

& \cellcolor{blue!5} w/ TTRL & \cellcolor{blue!5} 75.6\% & \cellcolor{blue!5} 88.9\% & \cellcolor{blue!5} 57.8\% & \cellcolor{blue!5} \textbf{23.3}\% & \cellcolor{blue!5} 38.4\% & \cellcolor{blue!5} 56.8\% \\
& \cellcolor{blue!5} Rel. $\Delta_{\text{TTRL}/\text{Base}}$ & \cellcolor{blue!5} \textcolor{red}{+48.2\%} & \cellcolor{blue!5} \textcolor{red}{+14.3\%} & \cellcolor{blue!5} \textcolor{red}{+152.4\%} & \cellcolor{blue!5} \textcolor{red}{+606.0\%} & \cellcolor{blue!5} \textcolor{red}{+24.7\%} & \cellcolor{blue!5} \textcolor{red}{+52.7\%} \\

\cdashlinelr{2-8} 

& \cellcolor{green!5} w/ CG-TTRL & \cellcolor{green!5} \textbf{78.6}\% & \cellcolor{green!5} \textbf{97.2}\% & \cellcolor{green!5} \textbf{59.0}\% & \cellcolor{green!5} \textbf{23.3}\% & \cellcolor{green!5} \textbf{38.9}\% & \cellcolor{green!5} \textbf{59.4}\% \\
& \cellcolor{green!5} Rel. $\Delta_{\text{CG-TTRL}/\text{Base}}$ & \cellcolor{green!5} \textcolor{red}{+54.1\%} & \cellcolor{green!5} \textcolor{red}{+24.9\%} & \cellcolor{green!5} \textcolor{red}{+157.6\%} & \cellcolor{green!5} \textcolor{red}{+606.0\%} & \cellcolor{green!5} \textcolor{red}{+26.3\%} & \cellcolor{green!5} \textcolor{red}{+59.7\%} \\

\midrule

\parbox[t]{2mm}{\multirow{18}{*}{\rotatebox{90}{\textbf{TTT on AMC}}}} 
& Qwen2.5-Math-1.5B-Instruct & 72.8\% & 86.1\% & 48.2\% & 10.0\% & 27.3\% & 48.9\% \\

& \cellcolor{blue!5} w/ TTRL & \cellcolor{blue!5} 75.4\% & \cellcolor{blue!5} 86.1\% & \cellcolor{blue!5} \textbf{54.2}\% & \cellcolor{blue!5} 16.7\% & \cellcolor{blue!5} 33.8\% & \cellcolor{blue!5} 53.2\% \\
& \cellcolor{blue!5} Rel. $\Delta_{\text{TTRL}/\text{Base}}$ & \cellcolor{blue!5} \textcolor{red}{+3.6\%} & \cellcolor{blue!5} \textcolor{red}{+0.0\%} & \cellcolor{blue!5} \textcolor{red}{+12.4\%} & \cellcolor{blue!5} \textcolor{red}{+67.0\%} & \cellcolor{blue!5} \textcolor{red}{+23.8\%} & \cellcolor{blue!5} \textcolor{red}{+8.8\%} \\

\cdashlinelr{2-8} 

& \cellcolor{green!5} w/ CG-TTRL & \cellcolor{green!5} \textbf{76.0}\% & \cellcolor{green!5} \textbf{91.7}\% & \cellcolor{green!5} \textbf{54.2}\% & \cellcolor{green!5} \textbf{20.0}\% & \cellcolor{green!5} \textbf{35.4}\% & \cellcolor{green!5} \textbf{55.5}\% \\
& \cellcolor{green!5} Rel. $\Delta_{\text{CG-TTRL}/\text{Base}}$ & \cellcolor{green!5} \textcolor{red}{+4.4\%} & \cellcolor{green!5} \textcolor{red}{+6.5\%} & \cellcolor{green!5} \textcolor{red}{+12.4\%} & \cellcolor{green!5} \textcolor{red}{+100.0\%} & \cellcolor{green!5} \textcolor{red}{+29.7\%} & \cellcolor{green!5} \textcolor{red}{+13.5\%} \\

\cmidrule{2-8} 

& Qwen2.5-1.5B-Instruct & 54.6\% & 72.2\% & 21.7\% & 3.3\% & 28.3\% & 36.0\% \\

& \cellcolor{blue!5} w/ TTRL & \cellcolor{blue!5} 56.4\% & \cellcolor{blue!5} 77.8\% & \cellcolor{blue!5} 30.1\% & \cellcolor{blue!5} \textbf{10.0}\% & \cellcolor{blue!5} 30.8\% & \cellcolor{blue!5} 41.0\% \\
& \cellcolor{blue!5} Rel. $\Delta_{\text{TTRL}/\text{Base}}$ & \cellcolor{blue!5} \textcolor{red}{+3.3\%} & \cellcolor{blue!5} \textcolor{red}{+7.8\%} & \cellcolor{blue!5} \textcolor{red}{+38.7\%} & \cellcolor{blue!5} \textcolor{red}{+203.0\%} & \cellcolor{blue!5} \textcolor{red}{+8.8\%} & \cellcolor{blue!5} \textcolor{red}{+13.9\%} \\

\cdashlinelr{2-8}

& \cellcolor{green!5} w/ CG-TTRL & \cellcolor{green!5} \textbf{56.8}\% & \cellcolor{green!5} \textbf{83.3}\% & \cellcolor{green!5} \textbf{35.0}\% & \cellcolor{green!5} \textbf{10.0}\% & \cellcolor{green!5} \textbf{32.3}\% & \cellcolor{green!5} \textbf{43.5}\% \\
& \cellcolor{green!5} Rel. $\Delta_{\text{CG-TTRL}/\text{Base}}$ & \cellcolor{green!5} \textcolor{red}{+4.0\%} & \cellcolor{green!5} \textcolor{red}{+15.4\%} & \cellcolor{green!5} \textcolor{red}{+61.3\%} & \cellcolor{green!5} \textcolor{red}{+203.0\%} & \cellcolor{green!5} \textcolor{red}{+14.1\%} & \cellcolor{green!5} \textcolor{red}{+20.8\%} \\

\cmidrule{2-8} 

& DeepSeek-R1-Distill & 51.0\% & 77.8\% & 22.9\% & 3.3\% & 30.8\% & 37.2\% \\

& \cellcolor{blue!5} w/ TTRL & \cellcolor{blue!5} 67.4\% & \cellcolor{blue!5} \textbf{87.7}\% & \cellcolor{blue!5} 42.2\% & \cellcolor{blue!5} 16.7\% & \cellcolor{blue!5} 33.3\% & \cellcolor{blue!5} 49.5\% \\
& \cellcolor{blue!5} Rel. $\Delta_{\text{TTRL}/\text{Base}}$ & \cellcolor{blue!5} \textcolor{red}{+32.2\%} & \cellcolor{blue!5} \textcolor{red}{+12.7\%} & \cellcolor{blue!5} \textcolor{red}{+84.3\%} & \cellcolor{blue!5} \textcolor{red}{+406.0\%} & \cellcolor{blue!5} \textcolor{red}{+8.1\%} & \cellcolor{blue!5} \textcolor{red}{+33.1\%} \\

\cdashlinelr{2-8}

& \cellcolor{green!5} w/ CG-TTRL & \cellcolor{green!5} \textbf{71.1}\% & \cellcolor{green!5} \textbf{87.7}\% & \cellcolor{green!5} \textbf{47.0}\% & \cellcolor{green!5} \textbf{20.0}\% & \cellcolor{green!5} \textbf{34.9}\% & \cellcolor{green!5} \textbf{52.1}\% \\
& \cellcolor{green!5} Rel. $\Delta_{\text{CG-TTRL}/\text{Base}}$ & \cellcolor{green!5} \textcolor{red}{+39.4\%} & \cellcolor{green!5} \textcolor{red}{+12.7\%} & \cellcolor{green!5} \textcolor{red}{+105.2\%} & \cellcolor{green!5} \textcolor{red}{+506.1\%} & \cellcolor{green!5} \textcolor{red}{+13.3\%} & \cellcolor{green!5} \textcolor{red}{+40.1\%} \\
\bottomrule
\end{tabular}
}
\caption{Main results: comparison of zero-shot, TTRL and our Context-Guided TTRL (CG-TTRL) on each task. Top block: test-time training (TTT) on MATH-500. Bottom block: TTT on AMC.}
\label{tab:ttrl_results_main}

\end{table*}

\keypoint{Domain Shift Performance} The results in Table~\ref{tab:ttrl_results_main} also allow us to compare in-domain and out-of-domain performance. We observe our CG-TTRL solution is typically helpful in both cases and it can lead to clear improvements in the presence of domain shift, both larger (GPQA) and smaller (other mathematics reasoning benchmarks).

\keypoint{Comparison with In-context Learning (ICL)} We study the impact of ICL on the base model without TTRL in the top part of Table~\ref{tab:training_context_selection_strategies_comparison}. For consistency with our CG-TTRL, we select the examples in the same way, using TF-IDF features. We see that while ICL improves over the performance of the base model, its improvement on its own is not as large as that of TTRL.

\begin{table*}[ht]
\centering
\resizebox{\textwidth}{!}{%
\begin{tabular}{lcccccc}
\toprule
\textbf{Name} & \textbf{MATH-500} & \textbf{GSM-100} & \textbf{AMC} & \textbf{AIME 2024} & \textbf{GPQA} & \textbf{Avg.} \\
\midrule
Base Model & 72.8\% & 86.1\% & 48.2\% & 10.0\% & 27.3\% & 48.9\% \\

\midrule

\cellcolor{orange!5}w/ In-context Learning & \cellcolor{orange!5} 72.3\% & \cellcolor{orange!5} 91.7\% & \cellcolor{orange!5} 50.6\% & \cellcolor{orange!5} 13.3\% & \cellcolor{orange!5} 28.3\% & \cellcolor{orange!5} 51.2\% \\
\cellcolor{orange!5} Rel. $\Delta_{\text{ICL}/\text{Base}}$ & \cellcolor{orange!5} \textcolor{red}{-0.7\%} & \cellcolor{orange!5} \textcolor{red}{+6.6\%} & \cellcolor{orange!5} \textcolor{red}{+5.0\%} & \cellcolor{orange!5} \textcolor{red}{+33.3\%} & \cellcolor{orange!5} \textcolor{red}{+3.7\%} & \cellcolor{orange!5} \textcolor{red}{+4.7\%} \\

\midrule 
\cellcolor{blue!5}w/ TTRL & \cellcolor{blue!5} 78.6\% & \cellcolor{blue!5} 91.7\% & \cellcolor{blue!5} 54.2\% & \cellcolor{blue!5} 20.0\% & \cellcolor{blue!5} 37.4\% & \cellcolor{blue!5} 56.4\% \\
\cellcolor{blue!5}Rel. $\Delta_{\text{TTRL}/\text{Base}}$ & \cellcolor{blue!5} \textcolor{red}{+8.0\%} & \cellcolor{blue!5} \textcolor{red}{+6.5\%} & \cellcolor{blue!5} \textcolor{red}{+12.4\%} & \cellcolor{blue!5} \textcolor{red}{+100\%} & \cellcolor{blue!5} \textcolor{red}{+37.0\%} & \cellcolor{blue!5} \textcolor{red}{+15.3\%} \\

\midrule


\rowcolor{green!5}
w/ CG-TTRL$_{\text{Random}}$ & 78.0\% & \textbf{94.4}\% & 55.4\% & 20.0\% & 34.8\% & 56.5\% \\
\rowcolor{green!5}
Rel. $\Delta_{\text{Random}/\text{Base}}$ & \textcolor{red}{+7.1\%} & \textcolor{red}{+9.6\%} & \textcolor{red}{+14.9\%} & \textcolor{red}{+100.0\%} & \textcolor{red}{+27.5\%} & \textcolor{red}{+15.5\%} \\

\cdashlinelr{1-7} 


\rowcolor{green!5}
w/ CG-TTRL$_{\text{SBERT}}$ & 78.0\% & \textbf{94.4}\% & 54.2\% & 20.0\% & 35.4\% & 56.4\% \\
\rowcolor{green!5}
Rel. $\Delta_{\text{SBERT}/\text{Base}}$ & \textcolor{red}{+7.1\%} & \textcolor{red}{+9.6\%} & \textcolor{red}{+12.4\%} & \textcolor{red}{+100.0\%} & \textcolor{red}{+29.7\%} & \textcolor{red}{+15.3\%} \\


\cdashlinelr{1-7} 

\rowcolor{green!5}
w/ CG-TTRL$_{\text{Hybrid}}$ & 77.6\% & \textbf{94.4}\% & 56.6\% & \textbf{23.3}\% & 36.9\% & 57.8\% \\
\rowcolor{green!5}
Rel. $\Delta_{\text{Hybrid}/\text{Base}}$ & \textcolor{red}{+6.6\%} & \textcolor{red}{+9.6\%} & \textcolor{red}{+17.4\%} & \textcolor{red}{+133.3\%} & \textcolor{red}{+35.2\%} & \textcolor{red}{+18.2\%} \\

\cdashlinelr{1-7} 

\rowcolor{green!5}
w/ CG-TTRL$_{\text{TF-IDF + MMR}}$ & 79.0\% & \textbf{94.4}\% & \textbf{57.8}\% & \textbf{23.3}\% & 37.4\% & \textbf{58.4}\% \\
\rowcolor{green!5}
Rel. $\Delta_{\text{TF-IDF + MMR}/\text{Base}}$ & \textcolor{red}{+8.5\%} & \textcolor{red}{+9.6\%} & \textcolor{red}{+19.9\%} & \textcolor{red}{+133.3\%} & \textcolor{red}{+37.0\%} & \textcolor{red}{+19.4\%} \\

\midrule


\rowcolor{green!5}
w/ CG-TTRL$_{\text{TF-IDF}}$ & \textbf{79.2}\% & \textbf{94.4}\% & 56.6\% & \textbf{23.3}\% & \textbf{38.4}\% & \textbf{58.4}\% \\
\rowcolor{green!5}
Rel. $\Delta_{\text{TF-IDF}/\text{Base}}$ & \textcolor{red}{+8.8\%} & \textcolor{red}{+9.6\%} & \textcolor{red}{+17.4\%} & \textcolor{red}{+133.3\%} & \textcolor{red}{+40.7\%} & \textcolor{red}{+19.4\%} \\

\bottomrule
\end{tabular}}
\caption{Comparison with in-context learning (ICL) and various context-selection strategies: Random, Sentence Bert (SBERT), hybrid, TF-IDF + MMR, and TF-IDF that we normally use as part of CG-TTRL. TTT on MATH-500 with Qwen2.5-Math-1.5B-Instruct model.}
\label{tab:training_context_selection_strategies_comparison}
\end{table*}

\begin{table}[h]
\centering
\resizebox{\columnwidth}{!}{%
\begin{tabular}{lcc}
\toprule
& \textbf{TTRL} & \textbf{CG-TTRL}$_{\text{\tiny TF-IDF}}$ \\
\midrule
\# Generated tokens & 760.3  & 706.3 \\
\# Total tokens &  850.0 & 1,687.0 \\
Time per iter. & 0.53h & 0.58h \\
Convergence iters. & 73.7 & 67.8 \\
Convergence time & 39.28h & 39.22h \\
\bottomrule
\end{tabular}}
\caption{Efficiency analysis of CG-TTRL vs TTRL in terms of the number of generated tokens, total number of processed tokens, time per iteration, convergence iterations (iterations until near top performance), total time until near top performance. Average across three models reported. Key observation is that CG-TTRL converges faster than TTRL.}
\label{tab:efficiency_analysis}
\end{table}

\begin{table*}[ht]
\centering
\resizebox{\textwidth}{!}{%
\begin{tabular}{lcccccc}
\toprule
\textbf{Name} & \textbf{MATH-500} & \textbf{GSM-100} & \textbf{AMC} & \textbf{AIME 2024} & \textbf{GPQA} & \textbf{Avg.} \\

\midrule


Qwen2.5-Math-1.5B & 34.6\% & 72.2\% & 33.7\% & 13.3\% & 26.3\% & 36.0\% \\
\rowcolor{blue!5}
w/ TTRL & 74.2\% & 94.4\% & 51.8\% & 23.3\% & 35.4\% & 55.8\%\\

\rowcolor{blue!5}
Rel. $\Delta_{\text{TTRL}/\text{Base}}$ & \textcolor{red}{+114.5\%} & \textcolor{red}{+30.7\%} & \textcolor{red}{+53.7\%} & \textcolor{red}{+75.2\%} & \textcolor{red}{34.6\%} & \textcolor{red}{+55.0\%} \\

\cdashlinelr{1-7} 

\rowcolor{green!5}
w/ CG-TTRL & 77.4\% & 94.4\% & 51.8\% & 23.3\% & 34.8\% & 56.3\% \\

\rowcolor{green!5}
Rel. $\Delta_{\text{CG-TTRL}/\text{Base}}$ & \textcolor{red}{+123.7\%} & \textcolor{red}{+30.7\%} & \textcolor{red}{+53.7\%} & \textcolor{red}{+75.2\%} & \textcolor{red}{+32.3\%} & \textcolor{red}{+56.4\%} \\


\midrule


Qwen2.5-Math-1.5B-Instruct & 72.8\% & 86.1\% & 48.2\% & 10.0\% & 27.3\% & 48.9\% \\

\rowcolor{blue!5} 
w/ TTRL & 78.6\% & 91.7\% & 54.2\% & 20.0\% & 37.4\% & 56.4\% \\

\rowcolor{blue!5} 
Rel. $\Delta_{\text{TTRL}/\text{BaseTTRL}}$ & \textcolor{red}{+8.0\%} & \textcolor{red}{+6.5\%} & \textcolor{red}{+12.4\%} & \textcolor{red}{+100\%} & \textcolor{red}{+37.0\%} & \textcolor{red}{+15.3\%} \\

\cdashlinelr{1-7} 

\rowcolor{green!5} 
w/ CG-TTRL & 79.2\% & 94.4\% & 56.6\% & 23.3\% & 38.6\% & 58.4\% \\

\rowcolor{green!5} 
Rel. $\Delta_{\text{CG-TTRL}/\text{Base}}$ & \textcolor{red}{+8.8\%} & \textcolor{red}{+9.6\%} & \textcolor{red}{+17.4\%} & \textcolor{red}{+133.3\%} & \textcolor{red}{+41.4\%} & \textcolor{red}{+19.4\%} \\


\bottomrule
\end{tabular}
}
\caption{Impact of CG-TTRL and TTRL on models trained with and without instruction tuning.}
\label{tab:instruction_tuning_impact}

\end{table*}

\keypoint{Context Selection Strategies} We compare with random selection of the examples and a number of more advanced strategies to select the context. These methods are less efficient than our solution based on TF-IDF and they include the following options. 
1) Using Sentence BERT model to extract embeddings and then using these to find the most similar examples. 2) Hybrid solution that combines TF-IDF with Sentence BERT embeddings (TF-IDF provides weights for the Sentence BERT embeddings). 3) Combination of TF-IDF with Maximal Marginal Relevance (MMR) weights \citep{Juseon+25,KapuriyaKGB25}, where we iteratively add the most relevant examples into the context set. We report the results for these other strategies for context selection in Table~\ref{tab:training_context_selection_strategies_comparison}. The results indicate that 
Random selection and 
Sentence BERT embeddings lead to similar performance as vanilla TTRL, with hybrid solution leading to improvements but not as large as vanilla TF-IDF. Combination of TF-IDF features with MMR obtains the same performance as vanilla TF-IDF, suggesting that the advanced iterative selection may not be needed.

\begin{figure}[h] 
    \centering
    \includegraphics[width=\columnwidth]{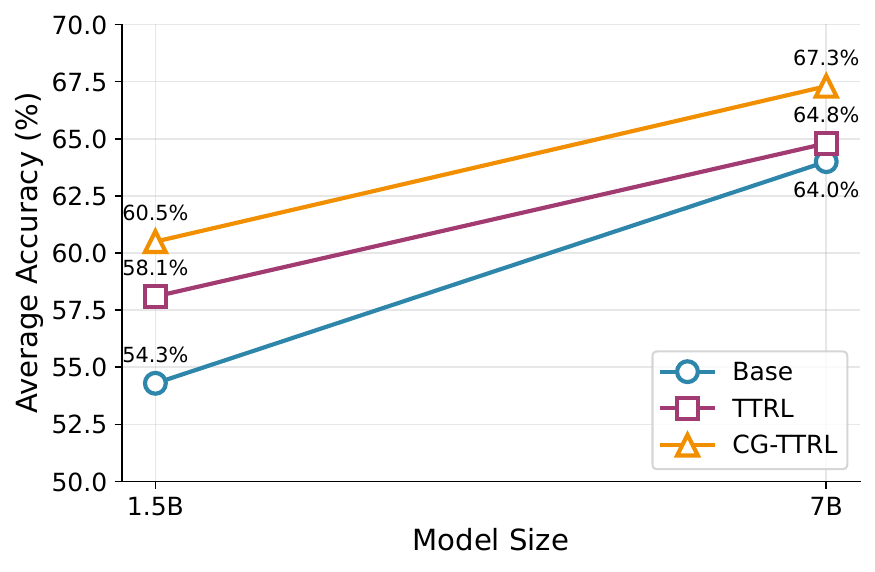}
    \caption{Model scaling analysis: our CG-TTRL method is useful also for larger models, e.g. ones with 7B parameters.}
    \label{fig:scaling_fig_avg}
\end{figure}

\keypoint{Model Scaling} While we focus mainly on on-device settings, we also tested the benefits of our method for larger models. In particular, we evaluate scaling of our method from Qwen2.5-Math-1.5B-Instruct model to its 7B variant. For these experiments we perform TTT on MATH-500 and evaluate across the different mathematical reasoning datasets. The results in Figure~\ref{fig:scaling_fig_avg} show we can successfully use our CG-TTRL method also for larger models. TTRL leads only to relatively minor improvements over the larger base model, but with our solution the improvements are noticeable.

\begin{figure}[h] 
    \centering
    \includegraphics[width=\columnwidth]{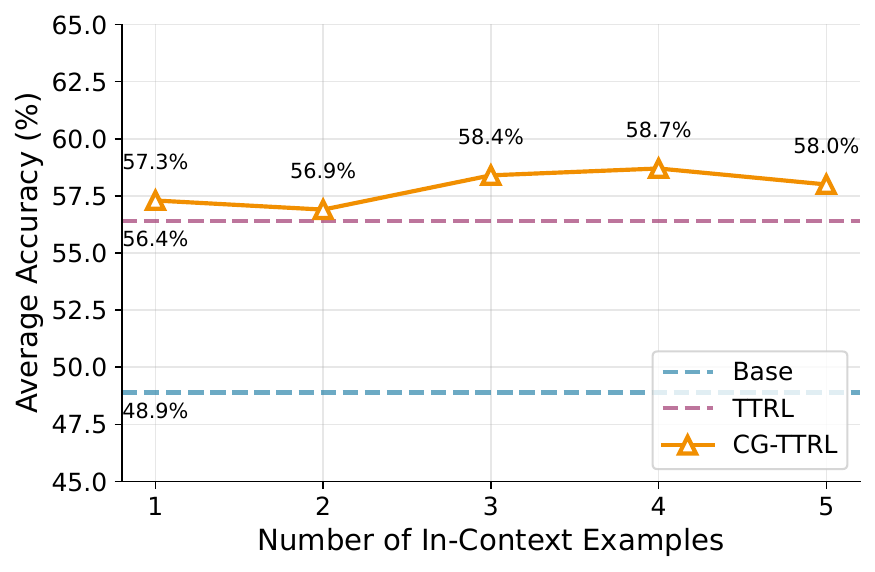}
    \caption{Number of context examples: using more examples is helpful in general, with consistent improvements of our CG-TTRL over both TTRL and the base model.}
    \label{fig:context_ablation_avg}
\end{figure}

\keypoint{Number of Context Examples} We study how the number of in-context examples influences the success of CG-TTRL in Figure~\ref{fig:context_ablation_avg}, on a scenario with TTT on MATH-500, evaluation across all considered datasets and using Qwen2.5-Math-1.5B-Instruct model. We see that using more examples in general helps improve the performance, but after some point adding more examples does not bring additional benefits.

\keypoint{Efficiency Analysis} We analyse the efficiency of our solution compared to vanilla TTRL in Table~\ref{tab:efficiency_analysis}. We consider a scenario where we use MATH-500 and report the average across the three models that we used for main evaluation. The results show that CG-TTRL leads to shorter generated outputs, even if the total number of processed tokens is larger due to the added context. The time per iteration is slightly longer for CG-TTRL, but it requires fewer iterations to reach near top performance. Overall it takes slightly less time for CG-TTRL to reach near top performance than for TTRL.

\begin{figure}[h]
    \centering
    \begin{subfigure}[b]{0.48\textwidth}
        \centering
        \includegraphics[width=\textwidth]{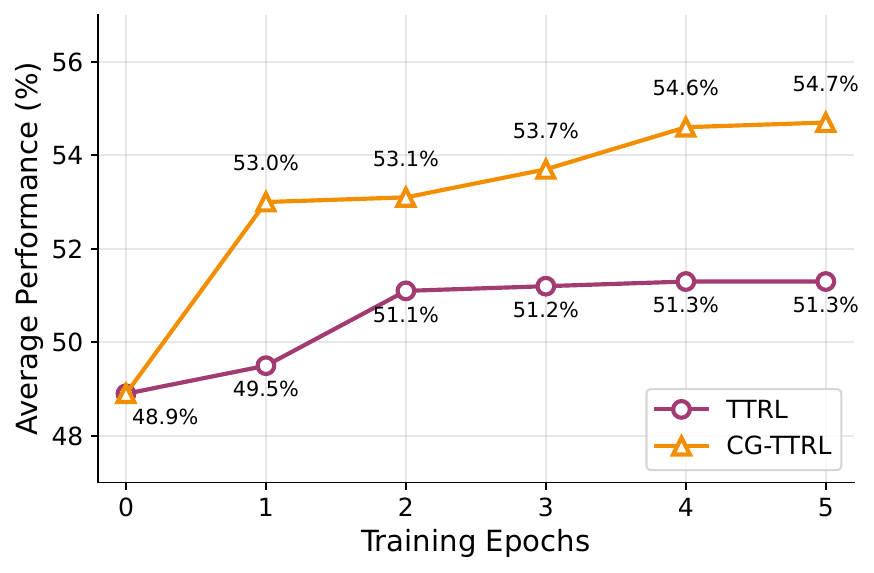}
        \label{fig:qwen_training}
    \end{subfigure}
    \caption{Few-epoch TTT analysis: CG-TTRL has especially strong improvements compared to TTRL when training only for very few epochs.}
    \label{fig:few_training_steps_comparison}
\end{figure}

\keypoint{Few-epoch Analysis} In on-device scenarios it may be more practical to adapt the model only using a small number of epochs (or steps), so we analyse the behaviour of both CG-TTRL and TTRL when using between 1 and 5 epochs for TTT. We use the MATH-500 dataset and Qwen2.5-Math-1.5B-Instruct model for this analysis, reporting the average result across all datasets. In this case one epoch corresponds to doing three update steps. Figure~\ref{fig:few_training_steps_comparison} shows CG-TTRL leads to a large increase even when using only one epoch for TTT (+8\% relative to zero-shot), unlike TTRL that improves more slowly (+1\% relative after one epoch). This observation highlights the usefulness of CG-TTRL for on-device use cases.

\begin{figure}[h!] 
    \centering
    \includegraphics[width=\columnwidth]{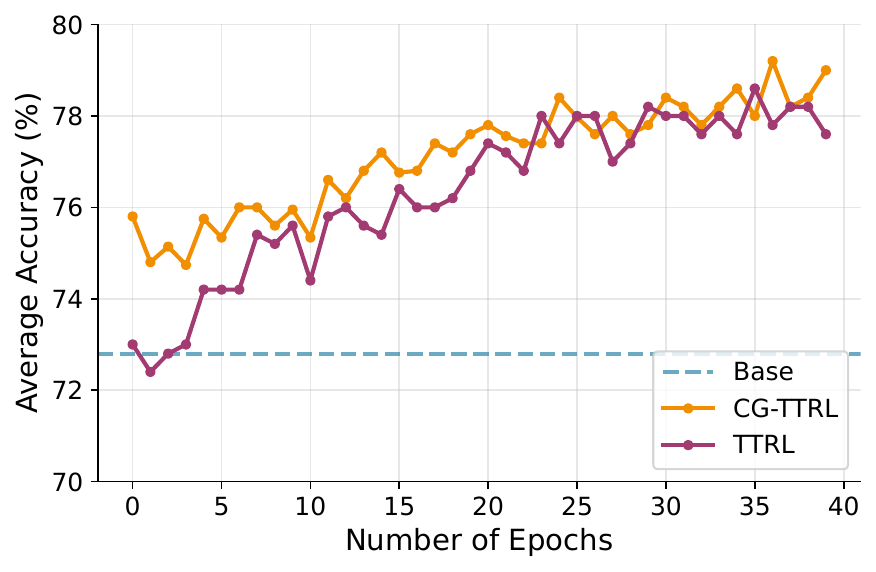}
    \caption{Training dynamics analysis: our CG-TTRL method leads to strong performance faster, but after significant amount of training the performance of TTRL becomes closer.}
    \label{fig:dynamics}
\end{figure}

\keypoint{Training Dynamics} We analyse the training dynamics in Figure~\ref{fig:dynamics}, using MATH-500 dataset and Qwen2.5-Math-1.5B-Instruct model. The analysis shows that CG-TTRL is able to obtain significantly stronger performance than TTRL when only a small number of epochs is used for training. After larger number of epochs the performance gap becomes smaller, with CG-TTRL still outperforming TTRL by a certain margin.

\keypoint{Instruction Tuning} We compare the impact of CG-TTRL on models trained with and without instruction tuning in Table~\ref{tab:instruction_tuning_impact}. The results show that CG-TTRL is useful for both, with stronger improvements seen for instruction tuned models that also perform better in general.

\section{Limitations}
\label{sec: Discussion}
In our paper we have focused on on-device scenarios and studied models that have suitable sizes for such use cases.
As shown on the analysis with a 7B model, our method can benefit also larger models and it is likely the benefits would extend to even larger models. However, TTRL is computationally expensive, so such analysis would be challenging to conduct with larger models. A limitation of TTRL more broadly is that it typically uses a larger number of rollouts, resulting in significant time required for the adaptation. In fact, the experiments take substantial time and so running TTRL until convergence is not practical in on-device scenarios. In such cases it is realistic to only perform e.g. one epoch of training.

\section{Conclusion}
\label{sec: Conclusion}
We have presented CG-TTRL technique that improves TTRL via efficiently selected context. Our solution has been designed to be especially suitable for on-device deployments, where efficiency and strong performance are crucial. Evaluation across a number of reasoning benchmarks has shown that CG-TTRL leads to noticeable improvements over TTRL's performance, while also obtaining strong performance improvement significantly faster. Relative improvements have been particularly large in the case of few-epoch adaptation where only a very small number of steps are used for the adaptation, a scenario especially suitable for on-device use cases.

%

%


\bibliography{tacl2021}

\end{document}